\documentclass[letterpaper]{article} 
\usepackage{aaai24}  
\usepackage{times}  
\usepackage{helvet}  
\usepackage{courier}  
\usepackage[hyphens]{url}  
\usepackage{graphicx} 
\usepackage{natbib}  
\usepackage{caption} 
\usepackage{algorithm}
\usepackage{algorithmic}

\usepackage{amsmath}
\usepackage{mismath}
\usepackage{amsfonts}
\usepackage{adjustbox}
\usepackage{booktabs}

\usepackage{newfloat}
\usepackage{listings}
\DeclareCaptionStyle{ruled}{labelfont=normalfont,labelsep=colon,strut=off} 
\floatstyle{ruled}
\newfloat{listing}{tb}{lst}{}
\floatname{listing}{Listing}
\title{Harnessing the Power of Beta Scoring in Deep Active Learning for Multi-Label Text Classification}
\author{
    Wei Tan\textsuperscript{\rm 1},
    Ngoc Dang Nguyen\textsuperscript{\rm 1},
        Lan Du\textsuperscript{\rm 1}\thanks{Corresponding author.},
        Wray Buntine\textsuperscript{\rm 2}
}
\affiliations{
    \textsuperscript{\rm 1}Faculty of IT, Monash University\\
    \textsuperscript{\rm 2}VinUniversity\\
     wei.tan2@monash.edu,
     dan.nguyen2@monash.edu,
     lan.du@monash.edu,
     wray.buntine@vinuni.edu.vn
}

\usepackage{bibentry}

\begin{document}

\maketitle

\begin{abstract}

Within the scope of natural language processing, the domain of multi-label text classification is uniquely challenging due to its expansive and uneven label distribution. The complexity deepens due to the demand for an extensive set of annotated data for training an advanced deep learning model, especially in specialized fields where the labeling task can be labor-intensive and often requires domain-specific knowledge. Addressing these challenges, our study introduces a novel deep active learning strategy, capitalizing on the Beta family of proper scoring rules within the Expected Loss Reduction framework. It computes the expected increase in scores using the Beta Scoring Rules,  which are then transformed into sample vector representations. These vector representations guide the diverse selection of informative samples, directly linking this process to the model's expected proper score.
Comprehensive evaluations across both synthetic and real datasets reveal our method's capability to often outperform established acquisition techniques in multi-label text classification, presenting encouraging outcomes across various architectural and dataset scenarios.
\end{abstract}

\section{Introduction}
\label{sec:Intro}

Text classification is an essential task in natural language processing (NLP) with various applications, including sentiment analysis and topic classification \cite{8719904, 9412588}. 
While deep learning has significantly improved multi-class text classification, yet its sub-task, multi-label text classification (MLTC), presents unique challenges. 
In MLTC, samples can be assigned multiple labels, navigating a vast, sparse label space with unresolved complexities \citep{10.1145/3077136.3080834, 8830456}. 
For every multi-label instance, the determination of appropriate labels adds to the intricacies and costs of annotation, making it even more resource-intensive than single-label learning. 
The increased demand for data becomes especially evident in specialized sectors, like the medical field, where domain-specific expertise and privacy considerations are crucial \citep{BUSTOS2020101797}.
Recently, active learning (AL) has been explored as a promising solution to MLTC's data scarcity issue, focusing on the annotation of the most informative instances. 
However, a comprehensive AL framework for MLTC, which robustly tackles imbalanced label distributions, has yet to be fully realized \citep{Cherman2019, 10.1145/3379504}. 
In spite of these challenges, MLTC remains a critical task in NLP with diverse practical applications.


AL can alleviated the complexities MLTC, primarily due to its capacity to identify high-information samples, which subsequently optimizes human annotation and enhances model performance \citep{REYES2018494, 10.1145/3379504}.
Binary Relevance (BR) plays a key role in MLTC for its ability to segment multi-label tasks into individual binary classifications, facilitating the inclusion of advanced classifiers like deep neural networks and support vector machines (SVMs) \cite{Zhang2018}. 
Inspired by BR's framework, earlier work in multi-label active learning (MLAL) adopted standard SVMs to establish binary classifiers tailored to individual labels \cite{mmc2009, adaptive2013, Cherman2019}.
Contrary to the notion that BR overlooks label correlation, when paired with deep neural networks, label correlations are implicitly handled, addressing a critical aspect of multi-label classification \cite{Su2021}. 
However, this focus on label correlation should not overshadow another crucial aspect of MLTC: data imbalance. The skewed distribution of labels in many real-world datasets affects instance selection strategies, potentially biasing models towards frequent labels \cite{Wu_Lyu_Ghanem_2016}. 
Therefore, recent studies, emphasize the need to explore the inherent properties of MLTC, especially label imbalance, as a pathway to enhance generalization performance, making it imperative to address this issue 
 precisely \cite{Zhang2018}.

In this paper, we propose a hybrid approach, named \textbf{Be}ta \textbf{S}coring \textbf{R}ules for Deep \textbf{A}ctive Learning (BESRA\footnote{Our implementation of BESRA can be downloaded from https://github.com/davidtw999/BESRA.}) that incorporates the Beta family of proper scoring rules within the Expected Loss Reduction (ELR) framework, applied to MLTC. 
Unlike expected loss used in other MLAL methods, BESRA pinpoints informative samples involving both uncertainty and diversity, drawing on expected score changes within a predictive model governed by the Beta scoring function.
Through extensive experimentation on diverse architectures ({\it e.g.,} TextCNN, TextRNN, BERT) and datasets ({\it e.g.,} Eurlex, RCV1, Bibtex, Delicious, Yahoo Health and TMC2007), we reveal that BESRA, even in the presence of varying label imbalances, consistently delivers robust performance, outperforming other AL methods. 
Our BESRA is adept handling of label imbalances sets a promising direction for future MLAL research and demonstrating robust performance in diverse data scenarios.

\section{Related Works}
\label{sec:Related}

AL is an iterative method that aims to reduce the cost and effort required for training high-quality prediction models by selecting the most informative instances from a pool of unlabeled data and querying an oracle for their labels \cite{settles2009active, BEMPS_Wei_NEURIPS2011}. From the literature, AL studies typically focus on multi-class classification (MCC) problems with two common strategies, including uncertainty-based and diversity-based sampling \cite{ren2020survey}. Uncertainty-based methods choose the instance that is most uncertain to label for the current trained classification model and quantify uncertainty using entropy or disagreement from ensembles \cite{holub2008entropy}. Diversity-based methods acquire informative and diverse samples using sample representations generated by pre-trained models \cite{ash2019deep}. However, relying solely on either uncertainty or diversity may not be adequate to select informative samples for annotation, as demonstrated by previous studies \cite{ren2020survey, TanDuBun-IEEEPAMI23}. To overcome this challenge, hybrid methods balance both uncertainty and diversity to acquire the most informative samples. These methods often generate sample representations from the loss that affects the model's performance and includes predictive uncertainty \cite{ash2019deep, TanDuBun-IEEEPAMI23}. 
Hybrid methods have shown promise in enhancing AL for MCC by combining the benefits of uncertainty and diversity.

AL for multi-label classification (MLC) is an area of ongoing research, and current methods often involve transforming the multi-label problem into one or more single-label problems, allowing for the application of traditional single-label learning algorithms \cite{Cherman2019, REYES2018494}. In MLC, Binary Relevance is a widely used method that decomposes the problem into a set of binary classification problems, enabling instance selection decisions through the independent exploitation of binary classifiers \cite{pmlr-v97-shi19b}.
For example, \citet{mmc2009} presents a strategy called maximum loss reduction with maximal confidence (MMC), which uses multi-class logistic regression to predict the number of labels for an unlabeled instance. Subsequently, it computes the expected
loss associated with the most
confident result of label prediction from support vector machine classifiers across all labels. 
To improve the MMC framework, \citet{pmlr-v20-hung11} proposed a soft Hamming loss reduction (SHLR) by extending MMC for MLAL by introducing a major and an auxiliary learner. The two learners are designed to evolve together during AL, and queries are made based on their disagreement. The framework provides flexibility in the choice of learners and query criteria and also includes MMC, BinMin, and the random query algorithm as special cases \cite{10.1007/978-3-319-97304-3_73, REYES2018494}.
Recently, ADAPTIVE \cite{adaptive2013} and CVIRS \cite{REYES2018494}  consider
label dependencies in their uncertainty sampling. 
The CSRPE \cite{10.1007/978-3-319-93034-3_12} method introduces cost-sensitive coding to manage varying error costs. Additionally, the GP-B2M \cite{gaupb2_2021} method deals with label sparsity and seeks to reveal label relationships using a Bayesian Bernoulli mixture of label clusters.
However, all these studies do not explicitly tackle the issues of label imbalance, which are frequently observed in MLTC, as indicated in Table \ref{tab:dataset-table}. 
Our AL method offers a fresh approach to addressing the imbalance challenges in MLTC, displaying consistent performance across both synthetic and real datasets.

\begin{figure*}[!t]
\centering
    \includegraphics[width=0.9\textwidth]{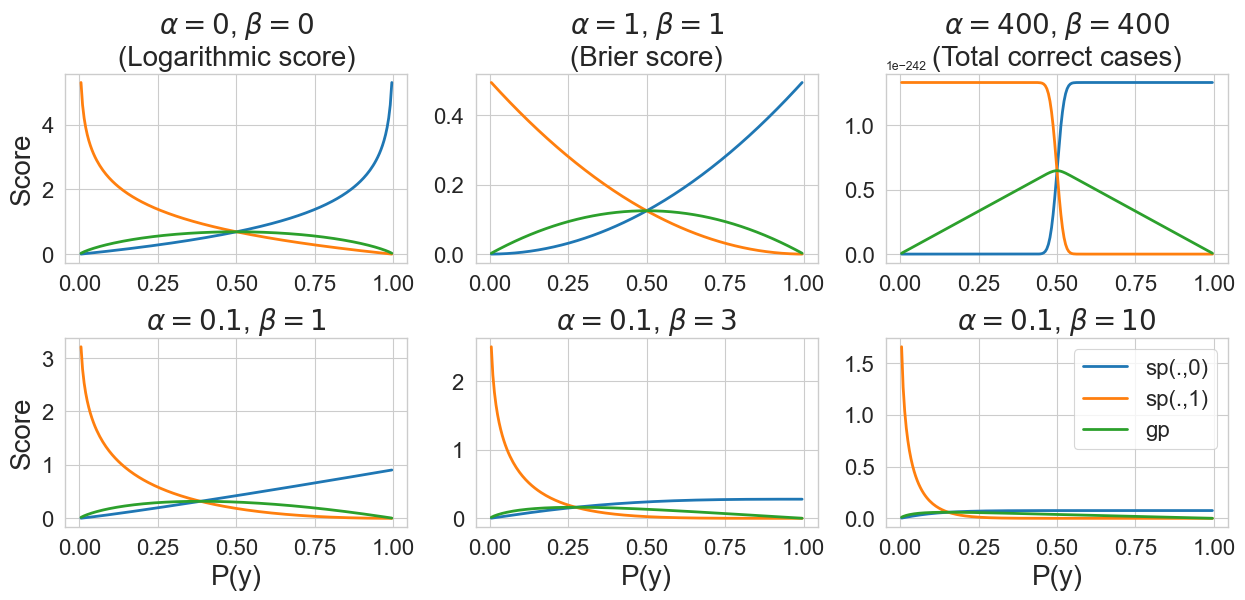}
    \caption{
    The graph depicts the Expected Score gp (green) and Scoring Functions from Eq~(\ref{eq-Sbeta}) for the Beta family in blue (sp(., 0) when $y=0$ and orange (sp(., 1) when $y=1$). It covers six scenarios: three specific to Brier Score, Logarithmic Score, and total error approximations, and three emphasizing asymmetry with varied Beta values.
    }
    \label{fig:betagp_blt}
\end{figure*}

\section{Beta Distribution-based Scoring Rules}
\label{sec:Method}

ELR \citep{roy2001toward} quantifies the generalisation error that is expected to reduce when a new sample is added to the labeled pool $L$. 
Leveraging the insights from previous research, it uses an unlabeled pool $U$ to represent potential test data, thereby quantifying the expected generalization error \citep{settles2009active}. 
Mean Objective Cost of Uncertainty (MOCU) \cite{zhao2020uncertainty} generalizes this approach and adds theoretical rigour.
The Bayesian Estimate of Mean Proper Scores (BEMPS) improves these formulations by replacing error with proper scoring rules \citep{doi:10.1198/016214506000001437}, leading to an AL strategy proven to converge  \citep{TanDuBun-IEEEPAMI23}.  
Scoring rules measure the quality of predictive distributions by rewarding calibrated predictive distributions.
Different scoring rules, including Brier score,
Logarithmic score, and Beta family \cite{doi:10.1198/016214506000001437},
focus the quality measurement on different aspects of the model prediction.
Unlike the Brier and Logarithmic scores, which are tailored for MCC, the Beta distribution inherently models variables limited between 0 and 1, making it more apt for BR in multi-label problems. 

The optimization goal of our multi-label active learner is to
label data that
leads to the largest change in expected scores
measured by the Beta family on the prediction of all binary classifiers. 
Let $P_T(x)$ be the (unknown) input distribution and
$P_T(y|x)$ be the (unknown) conditional distribution on label sets
$y\in \mathcal{Y}$, wherein $\mathcal{Y}$ is the set of all label sets.
This distribution $P_T(y|x)$ is used by the so-called Bayes classifier, an ideal because it is unknown.
We denote the predicted probability of $y|x$ given the current running model $\theta$ as $f_L(x,\theta)=P(y \mid \theta,x)$, 
and 
the predicted probability of $y|x$ under a Bayes optimal classifier
as $f_{L}(x)=P(y\mid L,x)=\int_\theta P(y \mid \theta,x) P(\theta|L) d\theta$. 

A ``quality"  score for the active learner is formulated
as the expected difference $Q_L$ 
between the score of the true probability estimates and a Bayes classifier after training on $L$ \citep{TanDuBun-IEEEPAMI23}, following for instance \citet[Sct.~4.1]{pmlr-v20-hung11}.
However, the  true conditional distribution $P_T(y\mid x)$ is unknown, so we make do with the current Bayesian estimate $f_L(x)$ and average the score differences across different models $\theta$:
{\scriptsize
\begin{eqnarray}
    Q_L 
    &=&  \mathbb{E}_{P_T(x)} 
    \mathbb{E}_{P(\theta|L)}  \Big[
    \mathbb{E}_{P(y|\theta,x)}  \nonumber\\
    && \big[
            S(f_L(x,\theta),y) - S(f_L(x),y)  
        \big] \Big] ~, \label{eq-QSS}\\
        \Delta{Q}(x|L)
        &=& Q_{L} - \mathbb{E}_{ P(y \mid L,x) }\big[ Q_{L+\{x,y\}}   \big]
        \nonumber\\
    &=&\mathbb{E}_{P(y\mid L, x)}
    \bigg[
        \mathbb{E}_{P(x') P (y' \mid L, (x, y), x')}
    \nonumber\\
    &&
        \big[
            S(P(\cdot \mid L, (x, y), x'), y')
           - S(P(\cdot \mid L, x'), y')
        \big]
    \bigg] \label{eq-DQS} 
\end{eqnarray}}
\noindent
Here $S$ is a scoring function which
scores a predictive probability distribution given the true label $y$. The active learner should evaluate Eq~\eqref{eq-QSS}- with $L$ given by ${L+ \{x,y\}}$ for each possible unlabeled data $x \in U$ to find the optimal query  $x^*$.
Since labels $y$ of any new sample of data $x$ are unknown,
the $Q_{L+ \{x,y\}}$ should be averaged over labels again using the Bayesian estimate.  The formula to maximise for unlabeled sample $x$ then becomes the difference, 
$\Delta{Q}(x|L)$, as in Eq~(\ref{eq-DQS}),
which can be further simplified \cite[Equation 6]{TanDuBun-IEEEPAMI23}.

 In the multi-label setting where there are $K$ classes, and each class can be made a binary classification problem with $y_{k} \in \{0,1 \}$, the scoring function using BR is
  $S_{BR}( f_L(x),y) =\sum_{k=1}^{K}S^k(f_L(x),y_k)$
where $S^k(\cdot,\cdot)$ is the score for each binary class \cite{dawid2014theory}.  Because of linearity, this works neatly with Eq~(\ref{eq-DQS}).


For implementation, we follow the algorithm of BEMPS in  \cite{TanDuBun-IEEEPAMI23} using our $S_{BR}$ score.
For this,
$\mathbb{E}_{P_T(x)}$ is approximated as the average over a smaller estimation pool (sampled from $U$). Moreover, $\mathbb{E}_{ P(\theta \mid L) }$ is approximated with an ensemble $\frac{1}{|\Theta|}\sum_{\theta\in\Theta} $
 as is 
$\mathbb{E}_{ P(y \mid L,x) }$ as
$\frac{1}{|\Theta|}\sum_{\theta\in\Theta} \mathbb{E}_{ P(y \mid \theta,x) }$.  The Bayesian update required for $P(y' \mid L+\{x,y\},x')$ is done by reweighting the ensemble:
{\scriptsize
\begin{eqnarray}
    P(y' \mid L, (x, y), x') &=& \sum_{\theta\in\Theta^E} P(y' \mid \theta, x') P(\theta \mid L, (x, y))
    \label{eq-3}\\
P(\theta \mid L, (x, y)) & \approx&
\frac{P(\theta \mid L) P(y|\theta,x)}
    {\sum_{\theta\in\Theta^E} P(\theta\mid L) P(y \mid \theta,x)} \label{eq-4}
\end{eqnarray} }

Now the sum of strictly proper scores leads to a strictly proper score \cite{dawid2014theory}.  So if $S^k(\cdot,\cdot)$ is a strictly proper score, then it follows that our  score $S_{BR}(\cdot,\cdot)$ will be strictly proper, and therefore all the theoretical benefits ensue for our MLAL framework, such as provable convergence.
We propose using the Beta family introduced by \cite{Buja2005Loss}, a two-parameter family of proper scoring rules
$\alpha,\beta > -1$ particularly useful for imbalanced classes or unequal costs.
{\scriptsize
\begin{eqnarray} 
\label{eq-Sbeta}
S^k_{\alpha,\beta}(p,y_k) = \left\{
  \begin{array}{lr}
    - \int_0^p c^\alpha (1-c)^{\beta-1} \mbox{d}c  \\ =
         - \frac{\Gamma(\alpha+1)\Gamma(\beta)}{\Gamma(\alpha+\beta+1)} I_p(\alpha+1,\beta)
          & , y_k=0\\
    - \int_p^1 c^{\alpha-1} (1-c)^{\beta} \mbox{d}c \\ 
    =
         - \frac{\Gamma(\alpha)\Gamma(\beta+1)}{\Gamma(\alpha+\beta+1)} I_{1-p}(\beta+1,\alpha) & ,y_k=1
\end{array} \right.
\end{eqnarray}}
\noindent
Here, $I_x(a,b)$ denotes the Incomplete Beta Function for $a,b>0$, 
and the probability vector $P(y)=(p, 1-p)$ given as $p$. 
If one of $a,b$ $\le 0$, an alternative evaluation is required. 

The Beta family of scoring rules generalise the Logarithmic score and the Brier score used in BEMPS as follows \cite{merkle2013choosing}:
When $\alpha=\beta=0$, Eq~\eqref{eq-Sbeta} reproduces the Logarithmic score
and when $\alpha=\beta=1$, it yields the Brier score.
From this perspective, BESRA can be regarded as the generalization of BEMPS.
When $\alpha = \beta \rightarrow \infty$, the scoring function becomes a step function, which approximates misclassification score. 
However, unlike the Brier and Logarithmic scores, which equally penalize both the low probability region (i.e., False Negatives (FN)) and the high probability region (i.e., False Positives (FP)) (as seen in Figure~\ref{fig:betagp_blt}), the Beta family allows for a more tailored approach. 
This adaptability is especially beneficial in scenarios with specific demands, such as imbalanced multi-label classifications (see Table~\ref{tab:dataset-table}).
Adjusting the $\alpha$ and $\beta$ values, the Beta Scoring Rules can be adjusted to differentially penalize false positives and negatives.
For example, by fixing $\alpha=0.1$ and tweaking $\beta$, we can incline the penalties more towards FN predictions.
Conversely, selecting the right $\beta$ and $\alpha$ is critical and varies based on the problem, as highlighted in \citep{merkle2013choosing, doi:10.1198/016214506000001437}. Our research explores how varying these parameter values for BESRA impacts MLAL's performance in the Beta Parameters section below.

\begin{algorithm}[tb]\small
\caption{Beta Scoring Rules For Deep Active Learning}
\label{alg:alg-ensemble}
  
\textbf{Require}: initial unlabeled pool $U$, initial labeled pool $L$, model ensemble $\Theta^E=\{\theta_1,...,\theta_E\}$ built from $L$ by retraining, pre-computed values for $P(y_k \mid \theta,x )$ and $P(y_k \mid L, x )$ for all relevant $y$ and $x$, acquire batch size $B$, acquire batch set $\mathcal{A}$, estimation pool $X$, number of acquisition iterations $N$, number of classes $K$.
\begin{algorithmic}[1]\small
\STATE Initialize: $i=0,L_0\leftarrow L, U_0\leftarrow U$
    
    \WHILE {$i<N$} 
   
    \STATE  $vec_{x,x'} = 0$
    \FOR{$x \in U_i$ , $x' \in X$}
        \FOR{$k \in K, y_k \in \{0, 1\}$}
    \STATE 
    Compute $P(\theta|L,(x,y_k))$ with Eq~\eqref{eq-4} for for each $y_k$
    \STATE 
    Compute $p(y'_k|L+\{x,y'_k\},x')$ with Eq~\eqref{eq-3} for each $y'_k$
     
    \ENDFOR
  \STATE $vec_{x,x'} +\!\!\!= 
    \Delta Q(x|L,x')$, where $\Delta Q(x|L,x')$ is computed by Eq~\eqref{eq-DQS} with
    the beta scoring function defined in Eq~\eqref{eq-Sbeta}.
    \ENDFOR
    \STATE $\mathcal{A} = \emptyset$
    \STATE $centroids$ = $k$-Means centers $(vec_{x\in U_i}, B)$
        
    \FOR{$c \in centroids$}
    \STATE $\mathcal{A}  ~\cup\!\!= \{$ argmin$_{x \in U_i}||c - vec_{x}|| \}$
    \ENDFOR

        \STATE $L_{i+1}\leftarrow L_i \cup \mathcal{A} $
        \STATE $U_{i+1}\leftarrow U_i \setminus \mathcal{A} $

    \ENDWHILE
  \end{algorithmic}
\end{algorithm}

After incorporating the obtained samples into the labeled set, we proceed to retrain the multi-label classifier (e.g., TextCNN, TextRNN and BERT) using randomly initialized parameters. This retraining occurs after each iteration of acquisition and is carried out using the binary cross-entropy loss (BCE), a commonly employed loss function in multi-label learning \cite{9319440}.
Specifically,
given a specific data point $x$, 
we compute the probability of $x$ being affiliated with class $k$,
$\forall k \in \{1, 2, \dots,  K\}$  as
$p(y_k = 1|x) = \frac{1}{1 + \exp(f_k(x))}$,
where $f_k$ is the Binary Relevance associated with class $k$.
BESRA's Algorithm~\ref{alg:alg-ensemble} uses Bayes theorem to estimate $P(\theta|L,(x,y_k))$ from $P(\theta|L)$, leveraging a predefined estimation pool $X$, a randomly selected subset from unlabeled pool, for estimating expected values $\mathbb{E}_{ P(x')}
[\cdot]$ in Eq~\eqref{eq-DQS}.
Each query is now represented as a vector of expected scored changes, i.e.,
$vec_{x,x'}$.
These vectors are then organized using k-Means clustering, and from each cluster, only the data points closest to the centroid are selected for precision and representation.

\section{Experiments}
\label{sec:experiment}

We evaluated BESRA's impact on synthetic and real-world datasets, particularly studying the influence of Beta Scoring Rules on different imbalance levels in synthetic data. Further, we benchmarked BESRA against state-of-the-art (SOTA) AL methods, affirming BESRA's efficacy in MLTC.


\begin{table}[!t]
  \centering
  \begin{tabular}{lllll}
    \toprule
  Dataset     & \#Document   & \#Vocabulary  & MeanIR \\
             & Train/Test   & /\#Label  & Train/Test \\
    \midrule
    RCV1-10     & 1,200/600 & 23,759/10  & 10/50 \\
    RCV1-50      & 1,200/600 & 23,117/10  & 50/50\\
   RCV1-200      & 1,200/600 & 21,074/10  & 200/50 \\
   RCV1-400      & 1,200/600 & 34,464/10 & 400/50 \\
    \bottomrule
  \end{tabular}
    \caption{Five datasets (including the test set) are created using the MeanIR values that represent the imbalance level of the multi-label datasets. Higher values suggest increased sparsity and imbalance.} 
  \label{tab:syndataset-table}
\end{table}
\begin{table}[!tb]
  \centering
  \begin{tabular}{lllllll}
    \toprule
            
    Dataset     & \#Document   & \#Vocabulary    & MeanIR \\
             & Train/Test   & /\#Label & Train/Test \\
    \midrule
      Bibtex     & 5,916/1,479 & 1,836/159  & 12/15 \\
    TMC2007   & 22,876/5,720 & 497/25  & 22/23 \\
      Delicious     & 12,872/3,219 & 501/982  & 72/73 \\
    RCV1     & 20,833/7,965 & 119,475/102  & 285/152 \\
    Eurlex  & 15,470/3,868 & 5,000/201  &  426/164 \\
    Yahoo   & 7,364/1,841 & 30,587/32  & 656/162 \\
   
    \bottomrule
  \end{tabular}
    \caption{Six benchmark datasets with their corresponding imbalance level MeanIR statistics. }
  \label{tab:dataset-table}
\end{table}

\begin{figure*}[!t]
\centering
\includegraphics[width=1\textwidth]{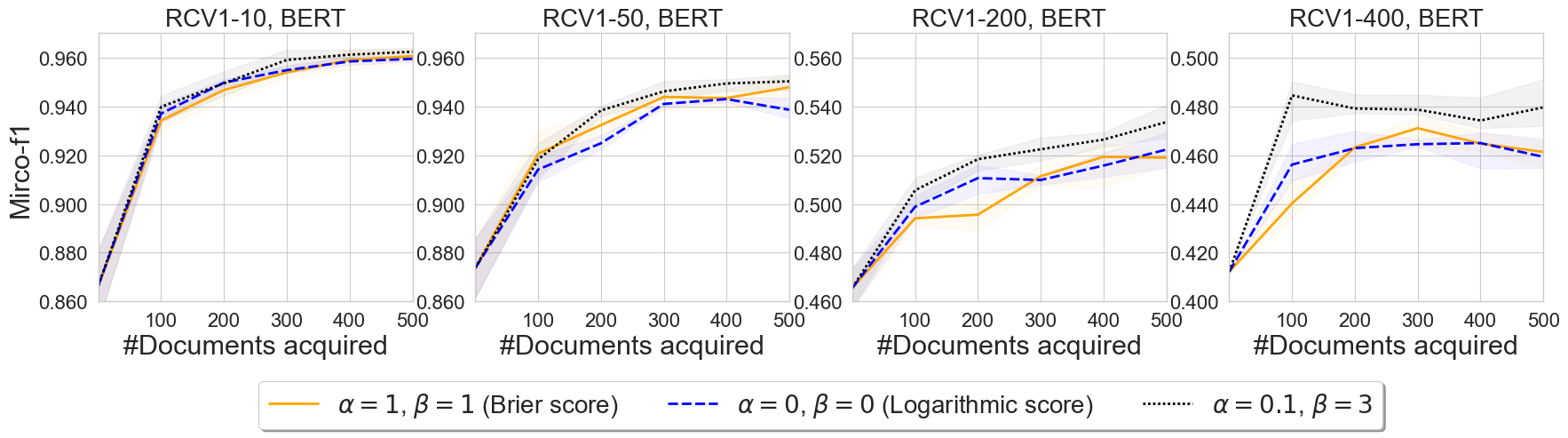}
    \caption{The average Micro F1-score of AL models with acquisition size 100 on BERT, which were run with 5 different random seeds on various synthetic datasets.}
\label{fig:syn_rcv1meanIR50_mircof1}
\vspace{-3mm}
\end{figure*}

\subsection{Datasets}

\subsubsection{Synthetic Datasets.}
We created five synthetic datasets, each consisting of ten labels and associated with a unique mean imbalance ratio (MeanIR), a measure of the average level of imbalance in the multi-label dataset \cite{CHARTE20153}, as outlined in Table~\ref{tab:syndataset-table}. 
Four of these datasets, aligned with MeanIR values of $(10, 50, 200, 400)$, served training purposes. Additionally, a consistent test set with a MeanIR of 50 was established for standardized evaluations \cite{wang2023imbalanced}.
To make the generation process for the synthetic dataset controllable, 
we reduced the label space to ten (i.e., $K=10$) by selecting the large dataset RCV1, and sampled instances based on the MeanIR metric provided by \citet{CHARTE20153}.
In contrast, label cardinality measures the average labels per instance and label density evaluates the instance-label proportion. However, neither offers a holistic perspective on the overall label imbalance.
As our main objective is to develop a general AL framework for MLTC that can handle the label imbalance issue,
These synthetic datasets are crucial for evaluating the performance of AL methods with the Beta Proper Scores under varying levels of label imbalance. 
The experiments used a greedy search to identify the optimal parameters for the Beta scoring function ($\alpha = 0.1, \beta = 3$) compared to the Brier score ($\alpha = 1, \beta = 1$) and the Logarithmic score ($\alpha = 0, \beta = 0$).
By systematically manipulating the Alpha and Beta values, the study evaluates BESRA's robustness and quantifies its influence across different magnitudes of dataset imbalance.

\subsubsection{Real Datasets.}

We have selected six diverse benchmark multi-label text datasets (MLTDs): Eulex \cite{LozaMencía2010}; RCV1 \cite{10.5555/1005332.1005345}, comprising Reuters newswire stories; Bibtex \cite{Katakis2008MultilabelTC}, which contains bibliographic records in BibTeX format; Delicious \cite{Tsoumakas2008EffectiveAE}, capturing user bookmarking data; Yahoo \cite{10.5555/2968618.2968710}, detailing health-related user discussions; and TMC2007 \cite{1559692}, a collection of texts from varied domains.
These datasets' label spaces vary significantly, ranging from 25 to 982 labels. We used the MeanIR statistics to measure the imbalance level in each. 
For example, Bibtex possesses the lowest MeanIR of the training set, meaning it's less imbalanced than the others. In contrast, Yahoo has the highest MeanIR, indicating it's the most imbalanced dataset among those we considered.
To ensure data integrity, we have kept both train and test sets in their original sizes. Table~\ref{tab:dataset-table} offers a summarized overview of these datasets' key characteristics for in-depth analysis and research.

\subsection{Model Architectures}

In our experiments, we use TextCNN \cite{kim-2014-convolutional}, TextRNN \cite{8632592}, and BERT \cite{devlin-etal-2019-bert} as the backbone classifier.
After each AL iteration, we fine-tune these models using random re-initialization \cite{frankle2018lottery} for enhanced efficacy—a method found to be more effective than incremental fine-tuning with new samples \cite{gal2017deep}. The fine-tuning process, executed on RTX3090 GPUs, sets a maximum sequence length of 256, runs for up to 80 epochs, and uses the AdamW optimizer with a 1e-5 learning rate \cite{nguyen-etal-2022-hardness,Nguyen_Tan_Du_Buntine_Beare_Chen_2023,nguyen2023lowresource}. Within the AL setup, the initial training and validation sets are composed of 100 and 1000 samples from the training set. To ensure robustness, each AL method iterates five times on each dataset with the random seeds, sampling 100 instances in each acquisition. By incorporating a deep ensembles method \cite{10.5555/3295222.3295387}, we compute predictive distributions $P(y| L, x)$ using five ensemble models. Performance was assessed using metrics like micro-F1, macro-F1, precision, recall, precision@5, and recall@5, though the primary paper emphasizes the micro F1-score, and further metrics' results can be found in the Appendix.

\subsection{Active Learning Baselines}

We compared BESRA with various AL methods, including a random baseline. 
For each method, we adapted its strategies to the MLTDs and using a consistent backbone classifier for all.
Among the methods examined, the \textbf{MMC} method by \citet{mmc2009} selects instances by maximizing loss reduction; \textbf{ADAPTIVE} by \citet{adaptive2013} chooses instances by balancing prediction uncertainty with label cardinality inconsistency; \textbf{AUDI} emphasizes label ranking and adjusted cardinality inconsistency, targeting uncertainty and diversity \cite{6729601}; \textbf{CVIRS} \cite{REYES2018494} leverages label vector inconsistency and score rankings to decide on instance labeling; \textbf{SHLR} by \citet{pmlr-v20-hung11} employs an auxiliary learner to determine instance choices based on disagreement levels; \textbf{CSRPE}, as introduced by \citet{10.1007/978-3-319-93034-3_12}, extends one-versus-one coding to a cost-sensitive approach, using code-bits to inform instance weights and improve performance; finally, \textbf{GPB2M} \cite{gaupb2_2021} combines a Gaussian Process with a Bayesian Bernoulli mixture to capture label correlations for instance acquisition.

\subsection{Results}

\subsubsection{Synthetic Datasets.}

Figure~\ref{fig:syn_rcv1meanIR50_mircof1} illustrates the micro F1-scores for three methods. Both the Brier ($\alpha=1$, $\beta=1$) and Logarithmic score ($\alpha=0$, $\beta=0$) equally penalize FNs and FPs. 
In contrast, the Beta score method with parameters ($\alpha=0.1$, $\beta=3$) emphasizes a more substantial penalty for FN predictions, leading to an enhanced AL performance.
Average results from five AL experiments, with a 95\% bootstrapped error band, underscore the Beta score's superior performance, especially on high imbalance datasets like RCV1-200 and RCV1-400. 
Although Brier shows a slight edge over the Logarithmic score in the RCV1-50 scenario, their differences diminish in other scenarios. 
Specifically, by optimizing for FN predictions, the Beta Proper Scores ($\alpha=0.1$ and $\beta=3$) have showcased their efficacy in handling varied imbalance distributions, especially as the imbalance ratio increases. 
However, these parameter values can be adjusted for specific applications based on desired scoring criteria, as discussed by \cite{Buja2005Loss, merkle2013choosing}. 
Further insights on this topic, including an ablation study and detailed discussion, are available in subsequent sections, with a more exhaustive analysis presented in the Appendix.

\begin{figure}[!t]
\centering
\includegraphics[width=0.47\textwidth]{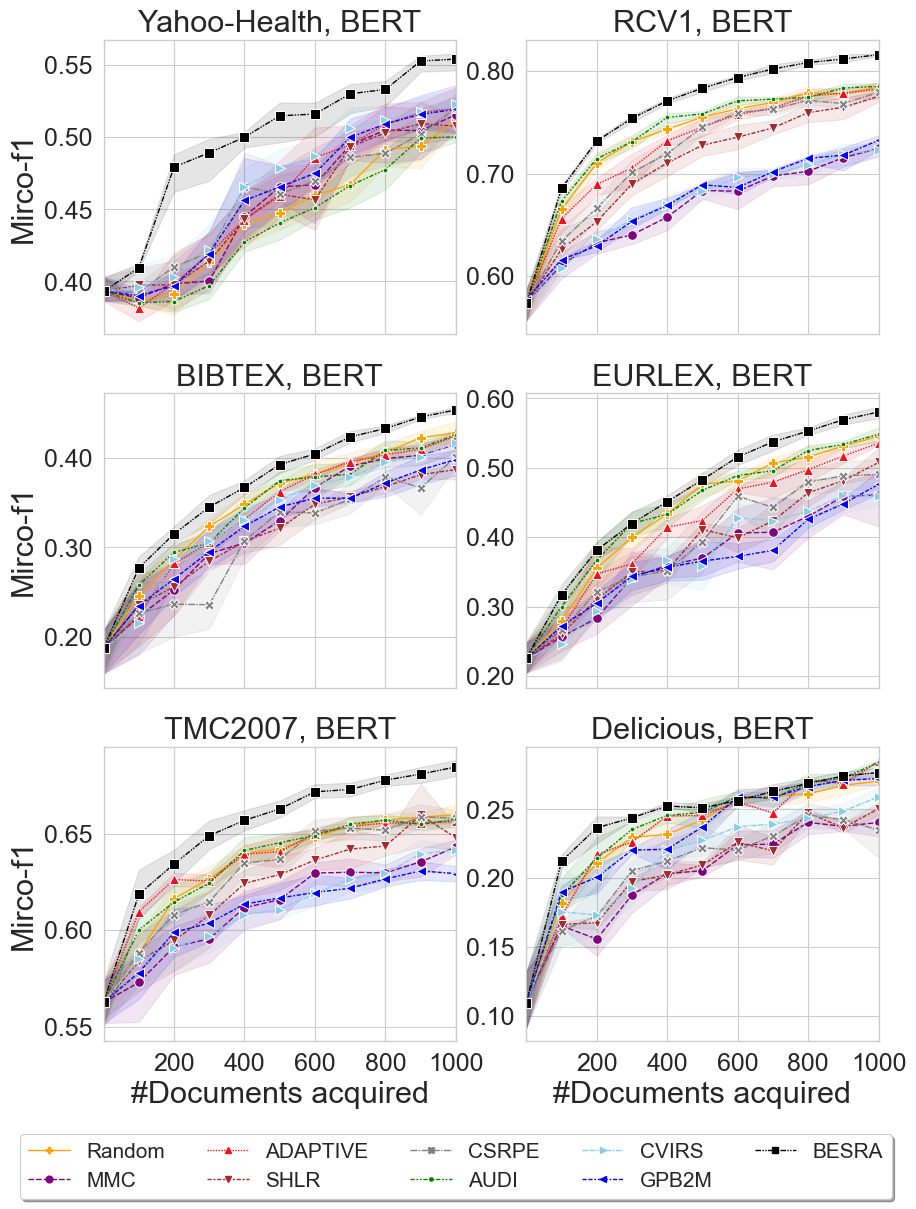}
    \caption{The average Micro F1-score of AL models with acquisition size 100 on BERT, which were run with 5 different random seeds on various datasets.}
    \label{fig:textBERT_mircof1_6}
    \vspace{-3mm}
\end{figure}

\subsubsection{Real dataset.} 
Figure~\ref{fig:textBERT_mircof1_6} shows a performance comparison between BESRA and the baseline methods across six MLTDs, as listed in Table~\ref{tab:dataset-table}.
We use the Beta score with $\alpha=0.1$ and $\beta=3$ due to its superior performance compared to the Brier score and Logarithmic score.
Remarkably, BESRA consistently outperforms other AL methods in all acquisition scenarios, regardless of the text domains. 
Among the evaluated methods, ADAPTIVE has the highest overall performance based on the micro F1-score. It consistently demonstrated strong performance across all datasets, showing its effectiveness in MLTC AL. 
Asides from ADAPTIVE, AUDI also provided competitive performance on all datasets except for Yahoo (health), suggesting its potential in specific domains. Conversely, CVIRS and GPB2M exhibited inconsistent performance across the datasets, struggling particularly in EURLEX, RCV1, and TMC2007. 
CSRPE method yielded inadequate results in the Bibtex and Delicious datasets, indicating a lack of effectiveness in those specific contexts. 
MMC emerged as the least effective method across all datasets except for Yahoo (health). In addition, MMC and SHRL showed similar performance, as both methods focused on measuring the expected loss of the model. It is worth highlighting that SHRL showcased better results than MMC specifically in the RCV1 and TMC2007 datasets, 
this could be due to SHLR being the improvement of MMC by introducing the soft Hamming loss \cite{pmlr-v20-hung11} as an alternative to the mean of the expected loss used in MMC.
GPB2M, the most recent work in ML AL, demonstrated superior performance in the Yahoo (health) and Delicious datasets, highlighting its efficacy in those specific contexts.
However, the robustness of this AL strategy raises a significant concern, as its performance exhibits variability not only across diverse datasets but also among various model architectures. 
For a more comprehensive analysis, including additional metrics such as macro-F1, precision and recall, please refer to the Appendix.

\begin{figure}[!t]
\centering
\includegraphics[width=0.47\textwidth]{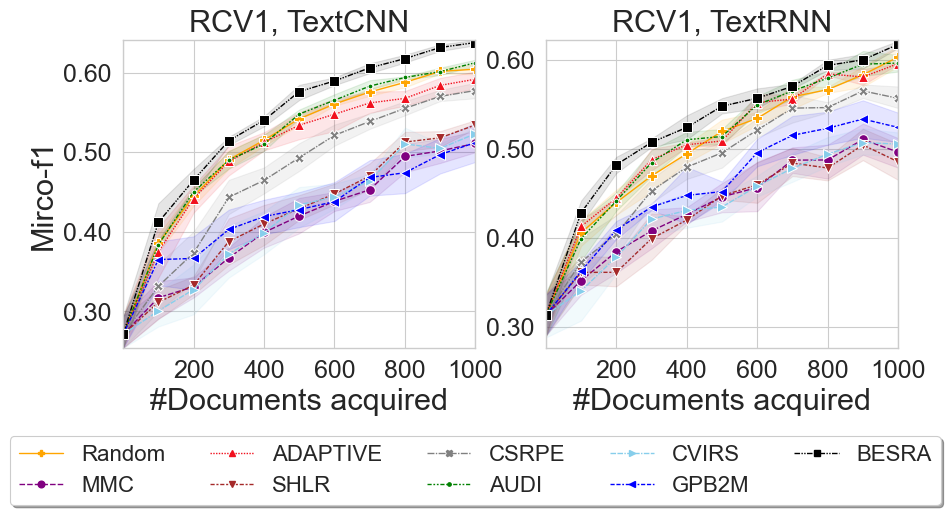}
    \caption{The average Micro F1-score of AL models with acquisition size 100 on TextCNN and TextRNN, which were run with 5 different random seeds on RCV1.}
    \label{fig:modelagnostic_mircof1_3}
     \vspace{-3mm}
\end{figure}

\section{Ablation Study}
\label{sec:ablation}

To assess the effectiveness and generalizability of BESRA, we conducted a comprehensive evaluation of three well-established neural network architectures commonly used in MLTC tasks: TextCNN, TextRNN, and BERT. These architectures have played a significant role in multilabel learning. 
Additionally, we investigated the impact of different Alpha and Beta values on the model's performance.
Our study aimed to answer two key research questions: \textbf{(i)} Does the performance advantage offered by BESRA generalise across diverse models and architectures?
\textbf{(ii)} How do the Alpha and Beta values influence the scoring mechanism and subsequently affect the model's performance across real-world datasets with the varied imbalance level? 
By addressing these questions, we aimed to gain insights of BESRA and its potential application in various MLTC tasks.

\begin{figure}[!t]
\centering
    \includegraphics[width=0.47\textwidth]{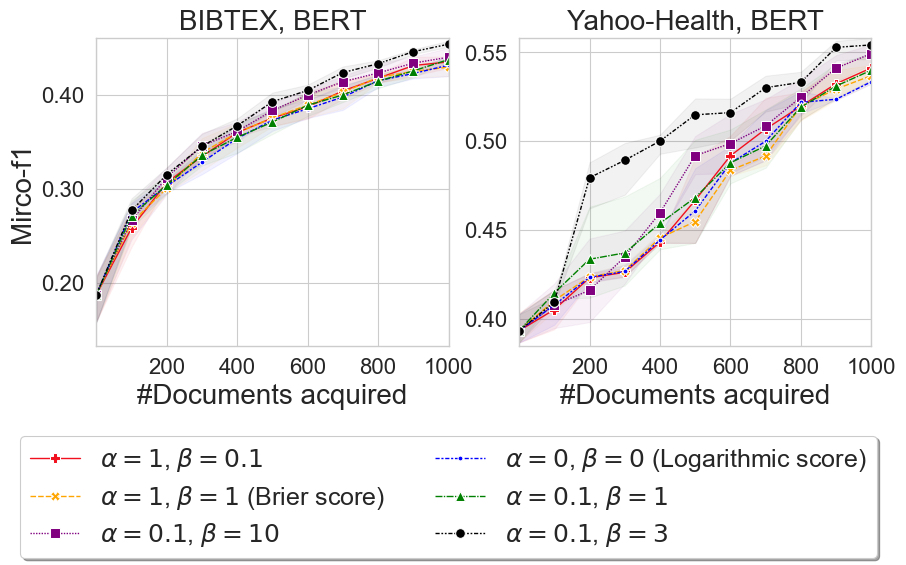}
    \caption{The average Micro F1-score of AL models with acquisition size 100 on BERT, which were run with 5 different random seeds on  for Bibtex and Yahoo (health).}
    \label{fig:ablationTextBERT_mircof1_6}
     \vspace{-3mm}
\end{figure}

\subsection{Model Generalizability}

The evaluation of BESRA is being conducted in two other models to test its generalizability.
As depicted in Figure \ref{fig:modelagnostic_mircof1_3}, BESRA consistently outperforms other baselines across TextCNN and TextRNN tested on the RCV1 dataset. Additionally, BESRA demonstrates exceptional performance compared to these models on five additional datasets included in the Appendix. These results validate that BESRA demonstrates effectiveness across a diverse range of pre-trained language models, highlighting its applicability irrespective of the specific model architecture applied.

\begin{figure}[!ht]
\centering
\includegraphics[width=0.47\textwidth]{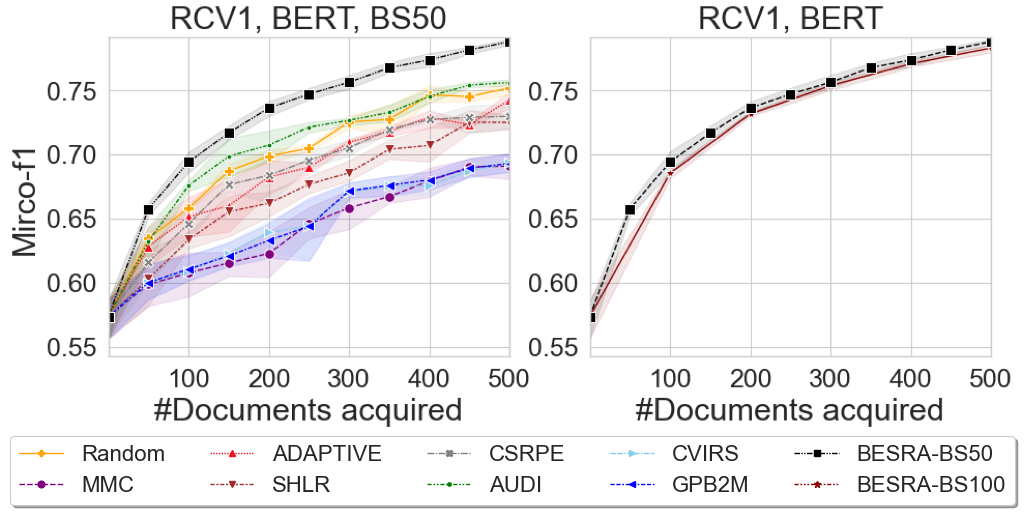}
    \caption{
    Left: Learning curves of ALs with batch size 50 on RCV1. Right: Learning curves for BESRA with batch sizes $B \in  \{50, 100\}$ on RCV1. All results were run with 5 different random seeds}
    \label{fig:batchsize_mircof1_2}
     \vspace{-3mm}
\end{figure}

\subsection{Batch Size}
Figure~\ref{fig:batchsize_mircof1_2} indicate that BESRA consistently outperforms other ALs when using a batch size of 50. 
Additionally, during the early stages of acquisition, BESRA performs more effectively with smaller batch sizes (50) than with larger batch sizes (100), aligning with the results reported in \cite{TanDuBun-IEEEPAMI23, NEURIPS2019_95323660}.
We hypothesise that at the early acquisition stages, due to the lack of knowledge, the model trained with a limited number of samples has high uncertainty and lacks calibration, thus acquiring a large batch of samples can result in noise. Additional experiments conducted on other MLTDs can be found in the Appendix.

\subsection{Beta Parameters}

In this subsection, we delve into the implications of varying Beta scores on the performance of AL, with a particular emphasis on how these scores influence the penalization behaviour of the active learner. 
We evaluated a range of $\alpha$ and $\beta$ values, associated with distinct Beta scores as shown in Eq~(\ref{eq-Sbeta}). 
Our evaluations spanned three representative datasets, ranging from relatively balanced (i.e., BIBTEX) to highly imbalanced (i.e., Yahoo). To gain a comprehensive understanding of how different $\alpha$ and $\beta$ values impact the outcomes, we considered several scoring methods including the Brier score and Logarithmic score (which provide equal penalization), alongside four distinct scenarios of Beta score. These scenarios include (1) mild penalization on False Positives (FPs) where $\alpha=1$ and $\beta=0.1$, (2) light penalization on False Negatives (FNs) with $\alpha=0.1$ and $\beta=1$, (3) moderate penalization on FNs defined by $\alpha=0.1$ and $\beta=3$, and finally (4) stringent penalization on FNs denoted by $\alpha=0.1$ and $\beta=10$.

Insights from Figure~\ref{fig:ablationTextBERT_mircof1_6} reveal that the Brier and Logarithmic scores, as well as the scenario with $\alpha=1$ and $\beta=0.1$, consistently underperform across various imbalance levels within MLTDs. 
Such results are consistent with our prior expectations, considering the inherent label imbalance challenges.  
A significant reason for this underperformance is the equal penalization rendered to both FN and FP outcomes by the Brier and Logarithmic scores. Additionally, the specific scenario of $\alpha=1$ and $\beta=0.1$ tends to disproportionately penalize FPs, thereby dampening performance. When evaluating the effect of $\alpha$ and $\beta$ values focused on penalizing FNs, we note that while light penalization settings (i.e., $\alpha=0.1$, $\beta=1$) have a negligible impact on enhancing the active learner's effectiveness, the more stringent configuration of $\alpha=0.1$ and $\beta=10$ offers notable improvements, especially in highly imbalanced datasets. However it does not necessarily culminate in the optimal active learner.
Instead, a moderate penalization strategy with $\alpha=0.1$ and $\beta=3$ consistently stands out as the most effective across MLTDs.

\section{Conclusion}

We have introduced BESRA, a novel acquisition strategy for MLAL.
This generalizes the recently published BEMPS using the Beta family of proper scoring rules,
which allow customizable asymmetric scoring rules
that effectively address the challenges such as imbalanced data associated with multi-label learning.
Moreover, by our methodical construction, the use of BESRA provably converges to optimal solutions.
Through empirical studies conducted on synthetic and real-world datasets, we have demonstrated  
the effectiveness of BESRA in acquiring highly informative samples for multi-label active learning, 
consistently surpassing seven existing acquisition strategies.   
This finding highlights the crucial role of Beta Scoring Rules and their great potential for AL with tailored acquisition strategies.
Future research can further explore combinations of Alpha and Beta values for specific datasets, addressing a current limitation of BESRA.

\bibliography{aaai24}

\end{document}